\documentclass{article} % For LaTeX2e
\usepackage{iclr2027_conference,times}

\usepackage{amsmath,amsfonts,bm}

\def\eqref#1{equation~\ref{#1}}
\def\1{\bm{1}}

\DeclareMathAlphabet{\mathsfit}{\encodingdefault}{\sfdefault}{m}{sl}
\SetMathAlphabet{\mathsfit}{bold}{\encodingdefault}{\sfdefault}{bx}{n}

\usepackage{hyperref}
\usepackage{url}

\title{SPDAlign: Interpretable Riemannian Alignment for EEG Forward Modeling Shifts}

\author{Shanglin Li$^{1,2,3,4}$, Shiwen Chu$^{3}$, Okan Ko\c{c}$^{4}$, Chenyu Liu$^{5}$, Qibin Zhao$^{4}$ \\ \textbf{Motoaki Kawanabe$^{3}$, Mitsuo Kawato$^{3}$ \& Yi Ding$^{5}$} \\
$^{1}$ BIFOLD, Germany, $^{2}$ Technical University of Berlin, Germany \\
$^{3}$ Advanced Telecommunications Research Institute International, Japan \\
$^{4}$ RIKEN AIP, Japan, $^{5}$ Nanyang Technological University, Singapore\\
}

\usepackage{booktabs} 
\usepackage{amsmath}
\usepackage{amsthm}
\theoremstyle{plain}

\theoremstyle{definition}

\theoremstyle{remark}

\usepackage{cleveref}
\crefname{equation}{Eq.}{Eqs.}
\Crefname{equation}{Equation}{Equations}
\crefname{figure}{Fig.}{Figs.}
\Crefname{figure}{Figure}{Figures}
\crefname{table}{Tab.}{Tabs.}
\Crefname{table}{Table}{Tables}
\crefname{section}{Sec.}{Secs.}
\Crefname{section}{Section}{Sections}
\crefname{appendix}{App.}{Apps.}
\Crefname{appendix}{Appendix}{Appendices}

\crefname{theorem}{Thm.}{Thms.}
\Crefname{theorem}{Theorem}{Theorems}
\crefname{lemma}{Lem.}{Lems.}
\Crefname{lemma}{Lemma}{Lemmas}
\crefname{definition}{Def.}{Defs.}
\Crefname{definition}{Definition}{Definitions}
\crefname{corollary}{Cor.}{Cors.}
\Crefname{corollary}{Corollary}{Corollaries}
\crefname{remark}{Rmk.}{Rmks.}
\Crefname{remark}{Remark}{Remarks}
\crefname{proposition}{Prop.}{Props.}
\Crefname{proposition}{Proposition}{Propositions}

\usepackage{pifont}
\newcommand{\cmark}{\ding{51}}% 勾
\newcommand{\xmark}{\ding{55}}% 叉
\usepackage{algorithm}
\usepackage{algorithmic}
\usepackage{graphicx} % 确保有这个包用于\scalebox

\usepackage{adjustbox}
\usepackage{color}
\usepackage[textsize=tiny]{todonotes}

\iclrfinalcopy % Uncomment for camera-ready version, but NOT for submission.
\begin{document}

\maketitle

\begin{abstract}
Electroencephalography (EEG) based brain-computer interfaces enable direct brain-to-device communication for applications such as rehabilitation and communication. 
However, their practical utility is often limited as the non-stationary nature of the EEG data introduces distribution shifts across domains (e.g., sessions and subjects).
Adapting machine learning models to be invariant to these shifts in an unsupervised way, without using costly labeled calibration data, would drastically improve the utility of EEG data.
In this work, we use a classic generative model of EEG to study distribution shifts introduced by the domain-specific forward process, which is associated with factors such as head geometry.
We theoretically show that such distribution shifts can be recovered solely through linear transformations on the Symmetric Positive Definite manifold.
Building on this insight, we propose SPDAlign, an interpretable framework for promoting domain-invariant EEG learning.
SPDAlign first aligns the domain-specific means and corrects global rotations across domains using a recent optimal transport technique called Wasserstein Procrustes. 
We systematically study the proposed approach through simulations and demonstrate its competitive performance on extensive public EEG datasets.
Additionally, SPDAlign is a globally linear framework and is intrinsically interpretable, so that the framework can identify frequency ranges of interest, determine the spatial patterns reflecting source–sensor relationships, and address cross-subject variability.

\end{abstract}

\section{Introduction}
Electroencephalography (EEG) records multi-channel brain activity from the scalp, providing noninvasive access to neuronal dynamics with millisecond resolution \citep{niedermeyer2005electroencephalography}.
EEG-based brain–computer interfaces (BCIs) enable direct brain-to-device communication for applications such as rehabilitation \citep{ang2016eeg}, emotion recognition \citep{lazarou2018eeg}, and assistive technology \citep{BCI}. 
Despite these promising applications, their practical utility is limited by poor generalization due to distribution shifts across subjects \citep{fairclough2020grand}. 

Conventional strategies for mitigating distribution shifts in EEG applications involve collecting labeled calibration data and training domain-specific models \citep{lotte2018review}. 
While effective in controlled settings, this approach poses significant challenges to the scalability of the technology \citep{wei_beetl_2022}.
As an alternative, domain adaptation aims to learn a model from source domains that generalizes effectively to unseen but related target domains, offering a principled alternative to training domain-specific models \citep{DAtheory}. 
Within the BCI field, domain adaptation mainly addresses the cross-subject transfer learning problem \citep{DAforBCI}, which is one of the grand challenges in EEG research \citep{fairclough2020grand}.

Unsupervised domain adaptation (UDA) \citep{wilson2020survey} without requiring target domain labels is considered the key to tackling this challenge \citep{wei_beetl_2022}.
Current literature casts it as a multi-source and multi-target UDA problem, attempting to address it with Riemannian alignment frameworks \citep{ju2025spd}. 
Operating on symmetric positive definite (SPD) manifolds, these alignment frameworks mitigate domain discrepancies by aligning first-order \citep{zanini2017transfer} and second-order \citep{kobler_controlling_2022} statistics, 
as well as rotational components \citep{rodrigues_riemannian_2019} across domains. 
Once aligned, a model trained on source domains typically generalizes to related target domains \citep{rodrigues_riemannian_2019}.
However, despite achieving state-of-the-art (SotA) performance \citep{lispdim, mellot2024geodesic}, these heuristic alignment methods lack theoretical foundations and thus offer limited explanatory power.

\begin{figure*}[t]
    \centering
    \includegraphics[width=\textwidth]{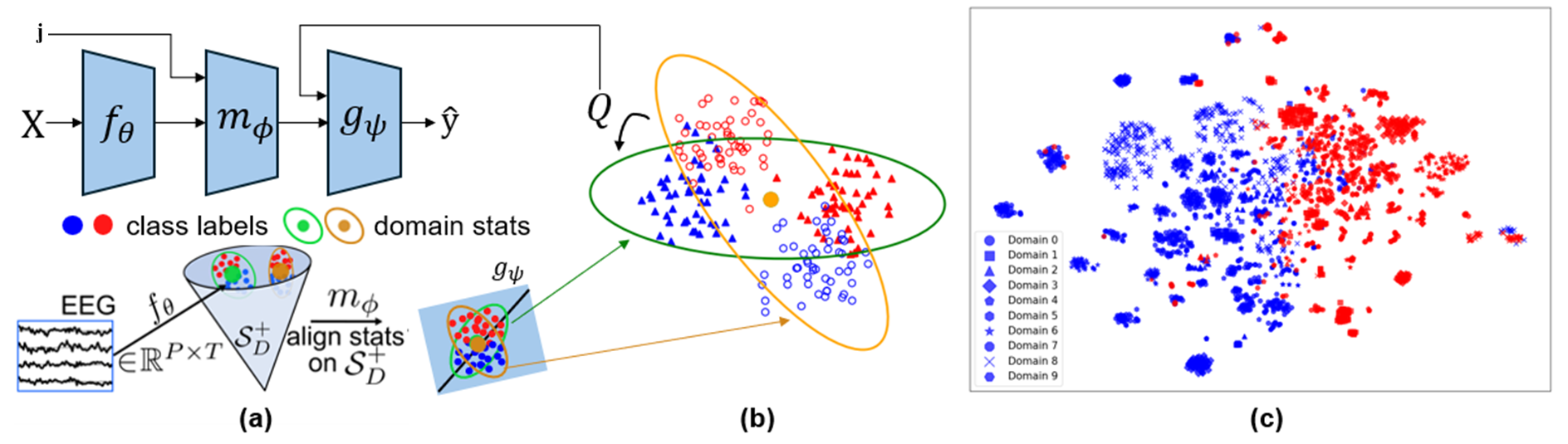}
    \caption{\textbf{SPDAlign Overview.}
    \textbf{(a) Architecture:} an end-to-end Riemannian learning framework on the SPD manifold that aligns domain-specific means and applies orthogonal transformations across domains for multi-source, multi-target unsupervised domain adaptation. 
    \textbf{(b) Rotational alignment:} Global rotation shifts between domains are resolved by estimating an orthogonal transformation \(Q\) via Wasserstein Procrustes. 
    \textbf{(c) t-SNE visualization:} Classification space for the driving fatigue dataset across all subjects. 
    Color represents labels, and shape denotes domains. 
    }
    \label{fig:overview}
\end{figure*}
In this work, we theoretically demonstrate that Riemannian alignment frameworks can compensate for distribution shifts introduced by the EEG forward modeling process in \cref{eq_forward_model}, which characterizes the observed EEG data as the linear mixing of latent source signals \citep{sabbagh2019manifold, haufe2014interpretation}.
This scenario represents a form of latent model shift, where discrepancies are driven by the domain-specific forward matrices associated with subject-specific variations such as head geometry \citep{mellot2023harmonizing, sabbagh2020predictive}.
By accounting for domain-specific forward matrices, we show that robust latent representations can be recovered solely through linear transformations on the SPD manifold.
Specifically, this involves mean matching followed by a global rotation transformation.
The resulting representations encode the latent source covariance, aligning with neuroscience studies that utilize source power as a robust biomarker \citep{blankertz2007optimizing, grosse2008multiclass}. 

Building on these theoretical insights and recent advances in end-to-end SPD learning \citep{kobler2022spd}, we propose SPDAlign (\cref{fig:overview} (a)), an end-to-end Riemannian learning framework designed to facilitate EEG domain-invariant representation learning.
SPDAlign aligns the domain-specific means (\cref{fig:overview} (a)) and subsequently applies a global rotation transformation across domains (\cref{fig:overview} (b)).
These rotations are estimated using orthogonal Procrustes analysis for labeled source domains and Wasserstein Procrustes for unlabeled target domains \citep{grave2019unsupervised}.
Unlike prior semi-supervised frameworks limited to single-source, single-target settings \citep{rodrigues_riemannian_2019, bleuze2022tangent}, SPDAlign generalizes to unsupervised multi-source, multi-target adaptation without requiring target domain labels. 
By addressing forward modeling shifts (\cref{fig:overview} (c)), it aligns representations across subjects and steps toward domain-invariant representations.

SPDAlign is a globally linear framework, comprising a linear feature extractor, linear transformations, and a linear classifier, and therefore is intrinsically interpretable.
Classifier weights can be mapped back to spatial patterns and source contributions, supporting transparent and clinically meaningful decision-making \citep{haufe2014interpretation}.
We systematically evaluate SPDAlign through simulations under the assumed generative model and show its competitive performance in EEG cross-subject decoding via extensive experiments on public datasets spanning motor imagery, workload, mental stress, and driving fatigue. 
Moreover, using the mental workload task as a case study, we show that SPDAlign successfully identifies task-discriminative source activities peaking in the theta band within the frontal and parietal regions in \cref{fig:inter_low}, which aligns with neurophysiological markers of mental workload \citep{borghini2014measuring, dehais2020neuroergonomics}.
These results further demonstrate that SPDAlign can identify frequency ranges of interest, determine the spatial patterns reflecting source–sensor relationships, and address cross-subject variability.

\section{Preliminaries}
\subsection{Multi-source multi-target UDA}
Let $x$, $y$, and $j$ denote random variables representing the features, associated true labels, and domains, respectively.
Given $M$ labeled source domain datasets $\mathcal{S} = \{(x_i, y_i, j_i) \mid x_i \in \mathcal{X},\, y_i \in \mathcal{Y},\, j_i \in \mathcal{J}_s\}_{i=1}^{L_s}$ and $N$ unlabeled target domain datasets $\mathcal{T} = \{(x_i, j_i) \mid x_i \in \mathcal{X},\, j_i \in \mathcal{J}_t\}_{i=1}^{L_t}$, where $\mathcal{X}$ denotes the input space, $\mathcal{Y}$ the label space, and $\mathcal{J}$ the set of domain identifiers.
Here, $L_s$ and $L_t$ denote the number of samples, and $y_i$ and $j_i$ represent the associated label and domain of sample $i$. 
In the multi-source multi-target UDA setting, we assume that all source and target domains share the same input and label spaces, and the objective is to learn a model $h$ from $\mathcal{S}$ that generalizes to unseen, unlabeled $\mathcal{T}$ via unsupervised domain adaptation.

\subsection{EEG forward modeling}
\label{sec_generative_model}

The EEG signals $x_i \in \mathcal{X} \subseteq \mathbb{R}^{P \times T}$, consisting of epochs of multivariate time-series with $T$ temporal samples and $P$ spatial channels, are modeled as a linear mixing of informative source plus additive noise \citep{nunez2006electric}:
\begin{equation}
    x_i = \mathbf{A}_s \mathbf{z}_i + \mathbf{n}_i,
    \label{eq_forward_model}
\end{equation}
where $z_i \in \mathbb{R}^{Q \times T}$ denotes the activity of informative sources and
$\mathbf{n}_i \in \mathbb{R}^{P}$ the additive noise. 
The matrix $\mathbf{A}_s \in \mathbb{R}^{P \times Q}$ contains the $Q$ source patterns.
As in \citet{sabbagh2020predictive, haufe2014interpretation}, the brain signals are assumed to arise from activity of uncorrelated sources, and the noise $\mathbf{n}_i = \mathbf{A}_n \boldsymbol{\nu}_i$ is stationary, uncorrelated with the sources, and spans a subspace
$\left(\mathbf{A}_n \in \mathbb{R}^{P \times (P-Q)}\right)$ that is shared across observations. 
The generative model can then be written as:
\begin{equation}
    x_i
    = A_s z_i
    + A_n \boldsymbol{\nu}_i
    = A_j s_i.
\end{equation}
where $s_i \in \mathbb{R}^{P \times T}$ denotes the activity of uncorrelated latent sources;
$A_j \in \{A \in \mathbb{R}^{P \times P }\}$ is a domain-specific forward model reflecting latent source activation patterns and is assumed to be invertible \citep{sabbagh2020predictive, kobler2021interpretation, mellot2023harmonizing}.

The task label $y_i$ is modeled as a function of the spatial covariance of the latent sources $E_i = \mathrm{Cov}(s_i)$:
\begin{equation}
    y_i = b^Tf(E_i)+ b_0 +\varepsilon_{i},
    \label{eq_linear_regression}
\end{equation}
where \( f(\cdot)\) is a fixed function, $b$ is a weight vector, $b_0$ is a bias term, and $\varepsilon_i \sim \mathcal{N}(0, \sigma^2)$ is the noise. 
Following the log-linear relationships typically observed in oscillatory brain activity \citep{buzsaki2014log}, 
we set \( f(\cdot) = \log(\cdot) \). 
Assuming zero-mean signals (i.e., $\sum_{t=1}^{T} x_t = 0$), the empirical covariance matrix for each multivariate epoch $x_i \in \mathcal{X}$ is defined as:
\begin{equation}
    C_{ij} = \mathrm{Cov}(x_i) = \frac{1}{T} x_i x_i^{\top} \in \mathcal{S}_P^{+}.
    \label{eq_sensor_covariance}
\end{equation}
Given the linear mixing between observed features and latent sources (\cref{eq_forward_model}), the sensor-space covariance matrices satisfy:
\begin{equation}
    C_{ij} = A_j E_i A_j^{\top}.
    \label{eq_model_C_i}
\end{equation}
Since $E_i$ encodes the target $y_i$ and $A_j$ is invertible, the empirical covariances $C_{ij}$ serve as sufficient statistics for decoding $y_i$. 

\subsection{SPD learning in EEG}
SPD-valued tangent space mapping (TSM) learning frameworks have been widely adopted in EEG analysis \citep{lotte2018review}. 
A nutshell introduction to SPD manifold is presented in \cref{app: spd_manifold}.
Their popularity stems from several favorable properties, including invariance to linear mixing \citep{congedo2017riemannian}, statistical consistency \citep{sabbagh2019manifold}, and interpretability \citep{kobler2021interpretation}.
In this context, the empirical covariance matrix $C_i$ (\cref{eq_sensor_covariance}) serves as a natural SPD representation \citep{barachant2011multiclass}. 
Typical TSM learning frameworks first estimate the Fr\'{e}chet mean $\bar{C}$ of the set $\mathcal{C} = \{\mathrm{Cov}(x_i) \mid x_i \in \mathcal{D}\}$ using \cref{eq_FM}, projecting each $C_i$ into the tangent space at $\bar{C}$ via \cref{eq_log_map}, and finally aligning the data around the identity matrix $I$ via parallel transport  \cref{eq_paratransport} for classification. 
Related work is discussed in \cref{app_rw}.

\subsection{Wasserstein Procrustes}
\textbf{Orthogonal Procrustes analysis.}
The orthogonal Procrustes problem seeks an orthogonal matrix \( Q \) that best aligns a given set of points \( X \in \mathbb{R}^{n \times d} \) with another set of points \( Y \in \mathbb{R}^{n \times d} \). 
When the correspondences between the two point sets are \emph{known} (i.e., which point of $X$ corresponds to which point of $Y$), the problem:
\begin{align}
Q^{\star} =: \arg\min_{Q \in \mathcal{O}_d} \; & \| XQ - Y \|_F^2, 
\label{eq:procrustes_objective} 
\end{align}
admits a closed-form solution \citep{schonemann1966generalized} via the singular value decomposition (SVD) of \( X^\top Y \). Let \( X^\top Y = U S V^\top \); then the optimal transformation is:
\begin{equation}
    Q^{\star} = U V^\top.
    \label{eq:procrustes_solution}
\end{equation}
In the above, \( \mathcal{O}_d \) denotes the set of \( d \times d \) orthogonal matrices and \( \|\cdot\|_F \) is the Frobenius norm.
The orthogonality constraint ensures that pairwise distances are preserved, restricting the alignment to a global rotation.

\textbf{Wasserstein distance.}  
When the correspondences between the two point sets $X$ and $Y$ are \emph{unknown}, the Wasserstein distance provides a principled way to infer correspondences by solving an optimal transport problem \citep{kuhn1955hungarian}. 
It finds a transport plan that distributes mass from points in $X$ to points in $Y$ in a way that minimizes the total transportation cost.
The squared 2-Wasserstein distance is defined as the distance between two discrete probability distributions  $\mathbf{X} = \{x_i\}_{i=1}^n$ with weights $a = (a_1, \dots, a_n)$ 
and $\mathbf{Y} = \{y_j\}_{j=1}^m$ with weights $b = (b_1, \dots, b_m)$ as:
\begin{equation}
W_2^2(X,Y)
\;=\;
\min_{\pi \in \Pi(a,b)}
\sum_{i=1}^n \sum_{j=1}^m
\pi_{ij} \, \|x_i - y_j\|_2^2,
\label{eq_wassterin_distance}
\end{equation}
where $\Pi(a,b)$ denotes the set of all transport plans 
$\pi \in \mathbb{R}_+^{n \times m}$ whose row sums equal $a$ and 
column sums equal $b$. 

\textbf{Wasserstein Procrustes.}
In our setting, neither the correspondence between the two point sets nor the underlying orthogonal transformation is known. 
Our objective is to learn an orthogonal matrix \( Q \in \mathcal{O}_d \) that aligns the set \( X \) with the set \( Y \), enabling the recovery of correspondences. 
To measure the similarity between the two sets, we employ the Wasserstein distance and integrate it with the orthogonal Procrustes formulation following \citep{grave2019unsupervised}. 
This yields the \emph{Wasserstein Procrustes} problem, where the goal is to estimate the orthogonal transformation $Q$ and the correspondences $\pi$ jointly:
\begin{equation}
\label{eq:wp}
\min_{Q \in O_d} 
W_2^2\!\left( X Q ,\, Y \right).
\end{equation}
Exact solutions exist for each sub-optimization problem when the other is fixed \citep{grave2019unsupervised}.

\section{Methods}

\subsection{EEG forward modeling shifts}
Following \citet{lispdim,sabbagh2020predictive}, we focus on EEG forward modeling shifts where the marginal distribution of the observed data varies across domains while assuming the conditional source-to-label relationship remains invariant. 
Under this assumption, a shared regression model in \cref{eq_linear_regression} can generalize effectively across domains once these distributional discrepancies are appropriately mitigated.
Specifically, we examine shifts introduced by the domain-specific forward matrix  $A_j$ (\cref{eq_forward_model}). 
This is a case of latent model shifting, where variations in the forward modeling process (head geometry, sensor placement, and cognitive states) induce distribution shifts in the observed sensor data \citep{mellot2023harmonizing}.
Changes in neural dynamics, such as impedance drift, may also occur but are typically modeled as noise \citep{akalin2013effects}.
By leveraging the uniqueness of the polar decomposition for square invertible matrices, we decompose $A_j$ as:
\begin{equation}
    A_j = Q_j \exp(P_j)
\label{eq_pd}
\end{equation}
where the transformation $A_j$ is expressed as a domain-specific transition $\mathrm{exp}(P_j)$ and orthogonal matrix $Q_j \in \{ Q \in \mathbb{R}^{P \times P} : Q^T Q = I_P \}$. 
We select the polar decomposition to characterize these shifts because it allows decomposing the possible shifts to translations and rotations on the SPD manifold, therefore naturally linking EEG forward modeling to common geometric operations on SPD matrices \citep{bhatia2019procrustes}.
Specifically, mean shifts correspond to global power changes that occur similarly across all channels \citep{congedo2017riemannian}, and rotation shifts correspond to scale shifts that differ across classes.

Let $\mathcal{C} = \{\mathrm{Cov}(x_i) \;|\; (x_i) \in \mathcal{D}\}$ denote the set of empirical covariance matrices. 
We define the first stage of our alignment framework, $m_{\phi_1}$, as the combination of TSM and recentering (RCT, \cref{eq_paratransport}) to the identity matrix on the manifold.
This stage aligns the domain-specific first-order moments (Fr\'{e}chet means) $\bar{C}_j$ to the identity matrix for each domain $j$:
\begin{equation}
\begin{aligned}
    m_{\phi_1}(C_{ij}) 
    &=   \Gamma_{\bar{C}_j \rightarrow I} \circ \mathrm{Log}_{\bar{C}_j}(C_{ij})
    =  \log\left( \bar{C}_j^{-\frac{1}{2}}\, C_{ij}\, \bar{C}_j^{-\frac{1}{2}} \right) \\
\end{aligned}
\label{eq_dsbn1}
\end{equation}

For the EEG forward modeling described in \cref{sec_generative_model}, an additional rotation transformation in \cref{eq_dsbn1} is beneficial to compensate for the distribution shifts induced by the domain-specific forward matrix $A$.
This transformation recovers robust latent representations $\log(E_i)$ that maintain a linear relationship to the labels $y_i$ via \cref{eq_linear_regression}:
\begin{equation}
   m_\phi(C_{ij}) = \hat{Q}_{j}^T \log \left( \bar{C}_{j}^{-\frac{1}{2}} C_{ij} \bar{C}_{j}^{-\frac{1}{2}} \right) \hat{Q}_{j}
    = \log(E_i)
    \label{eq_dsbn}
\end{equation}
In our framework, $\hat{Q}_{j}$ is estimated via orthogonal Procrustes analysis (\cref{eq:procrustes_solution}) using labels during training, and via Wasserstein Procrustes (\cref{eq:wp}) without labels during target inference. 
The outputs $m_{\phi}(C_i)$ are the robust latent representations and are forwarded to the regression model \cref{eq_linear_regression}.

\subsection{SPDAlign Framework}
\label{sec_decoding_approach}
As demonstrated by \cref{eq_dsbn}, robust latent representations $\log(E_i)$ can be recovered by compensating for distribution shifts introduced by the domain-specific forward matrix $A_j$ (\cref{eq_forward_model}) via alignment of domain-specific first-order moments followed by a global rotation.  
Building on these theoretical insights, we propose SPDAlign: an end-to-end Riemannian deep learning framework on the SPD manifold designed for multi-source/multi-target unsupervised domain adaptation.
SPDAlign addresses forward modeling shifts by leveraging source domain labels when available, while operating entirely without labels during target domain inference.

\begin{figure*}[t]
    \centering
    \includegraphics[width=\textwidth]{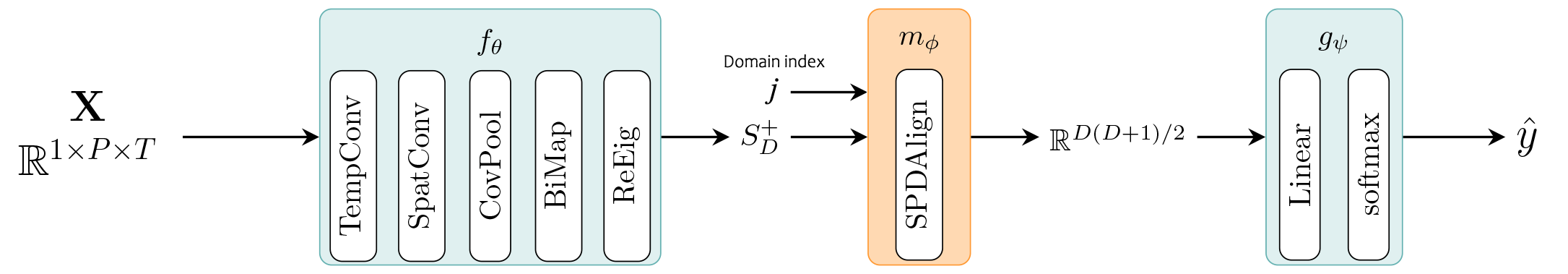}
    \caption{\textbf{SPDAlign architecture.} 
    SPDAlign optimizes parameters $\Theta = \{\theta, \phi, \psi\}$ in an end-to-end fashion: (i) a feature extractor $f_\theta$ maps raw EEG signals onto the SPD manifold $S^+_D$; 
    (ii) the SPDAlign module $m_\phi$ addresses shifts by aligning means and correcting rotations (\cref{eq_dsbn1,eq_dsbn}); 
    and (iii) a shared linear classifier $g_\psi$ outputs predictions $\hat{y}$. 
    By maintaining global linearity, SPDAlign is equivalent to the Tangent Space Mapping framework, providing interpretability.
    }

    \label{fig:SPDAlign}
\end{figure*}

\textbf{Source training.}
Following \citet{kobler2022spd}, we constrain the hypothesis class $\mathcal{H}$ to functions $h: \mathcal{X}\times \mathcal{D} \rightarrow \mathcal{Y}$ that can be decomposed into a composition of a shared feature extractor $f_{\theta}: \mathcal{X} \rightarrow \mathcal{S}_D^+$, an alignment module $m_{\phi}: \mathcal{S}_d^+ \times \mathcal{D} \rightarrow \mathbb{R}^{D(D+1)/2}$, and a shared linear classifier $g_{\psi}: \mathbb{R}^{D(D+1)/2} \rightarrow \mathcal{Y}$  with parameters $\Theta = \{\theta, \phi, \psi\}$. 
We parametrize $h=g_{\psi}\circ m_{\phi}\circ f_{\theta}$ as a neural network (details in \cref{sec_spdalign_architecture}) and learn the entire model in an end-to-end fashion, see \cref{fig:SPDAlign}. 
To tackle forward modeling shifts, $m_\phi$ tracks the running domain-specific Fr\'{e}chet mean during training and recenters these first-order moments to the identity matrix; and employs orthogonal Procrustes analysis in \cref{eq:procrustes_objective} to estimate the rotation between each domain data point and a reference domain, with labels providing known correspondences.
Finally, $g_{\psi}$ serves as the shared linear classifier.

Model training proceeds in two stages (\emph{\cref{alg:training}}). 
In the first stage (70\% of the training iterations), we train the model without applying the rotation alignment. 
This stage learns a good feature extractor that can extract common representations across domains. 
In the second stage, we freeze the parameters $\theta$ of the feature extractor, estimate and apply the domain-specific rotation alignment, and subsequently fine-tune the classifier parameters $\psi$ on the aligned representations.
To estimate the rotation transformation, the Orthogonal Procrustes (\cref{eq:procrustes_objective}) is applied during source training using labels to construct class-constrained sample correspondences. 
We select the reference domain randomly.
For each remaining domain, we compare its class-wise sample counts with those of the reference domain and, for any class with unequal sample counts, downsample the larger subset to match the smaller one.

SPDAlign builds on standard simplifying assumptions regarding the linear generative process of EEG \citep{sabbagh2019manifold, kobler2021interpretation}. 
In practice, the method is applied to covariance representations obtained after feature extraction. 
Because convolutions act as linear operators, this extraction step inherently preserves the structural form of the spatial generative model, so our theoretical analysis remains applicable \citep{haufe2014interpretation}.

\textbf{Target inference.}
During inference, SPDAlign freezes the learned feature extractor $f_{\theta}$ and the shared linear classifier $g_{\psi}$.  
When samples from previously unseen target domains arrive, it leverages the entire target domain data to estimate the domain-specific mean (\cref{eq_FM}) and the rotation transformation between the reference source domain (set in the training stage) and the target domain data via Wasserstein Procrustes (\cref{eq:wp}).  
In cases of label shifts, the Wasserstein component can be replaced with Unbalanced Optimal Transport (UOT) \citep{peyre2019computational, koccincorporating}, which relaxes marginal constraints.
The target rotations are estimated totally without target labels, and the resulting optimization procedure is detailed in \emph{\cref{alg:wp}}.
At each iteration $t$, we compute the optimal transport coupling between the entire reference domain and the target domain data.
The resulting transport plan is then used to compute the gradient with respect to the orthogonal matrix $\hat{Q}$, which is updated via projected gradient descent on the orthogonal group. 
We initialize $\hat{Q}$ to the identity matrix for simplicity and iterate for 50 epochs with a learning rate of 0.01.

\textbf{Interpretability.} 
\citet{kobler2021interpretation} demonstrated that the coupling of latent sources can be recovered by solving a generalized eigenvalue problem between the Fr\'{e}chet mean and classifier patterns \citep{haufe2014interpretation}.
With an additional linear orthogonal transformation, SPDAlign maintains global linearity, equivalent to the typical TSM learning framework.
This global linearity ensures the model preserves a direct correspondence between classifier weights and latent source contributions: the learned representations can be mapped back to the sensor space, and the resulting spatial patterns reflect source–sensor relationships and source contributions.
Such interpretability is crucial for determining the specific brain regions (measurement channels) and frequencies that reflect certain cognitive processes and for supporting trustworthy clinical BCI systems \citep{haufe2014interpretation}.

\section{Experiments}

To evaluate the effectiveness of SPDAlign, we simulated binary classification tasks using balanced accuracy as the evaluation metric, based on the forward model (implementation details in \cref{app_section_simulation}).
We compared the performance of SPDAlign against RCT \citep
{zanini2017transfer} across a range of rotational shifts.
We varied the parameters related to the forward model: class separability \cref{fig:main_rcc}, sample size \cref{fig:app_sam}, dimensionality \cref{fig:dim}, number of source domains \cref{fig:domain}, informative source \cref{fig:app_info}, data availability \cref{fig:app data percentage}, and label shifts \cref{fig:app label shift}.
We discuss the main findings here, and detailed results are provided in the Appendix \cref{app_section_simulation}.
As expected, classification performance declines as the magnitude of rotational shifts increases, and SPDAlign is superior to RCT across all scenarios.
The simulations varying parameters suggest that SPDAlign is robust within typical EEG regimes, where moderate sample sizes are available, low SNR, and sufficiently informative sources. 
These conditions are consistent with standard BCI datasets, where SPDAlign achieves strong performance. 

\begin{table*}[t]
\caption{\textbf{Cross-subject decoding results on four EEG tasks.}
Results are reported as the grand average of all test-set balanced accuracy (\%) $\pm$ standard deviation. 
The best performing model for each dataset is highlighted. 
Note that SPDAlign (Oracle) serves as an empirical upper bound, using target domain labels only for optimal rotation estimation through empirical risk minimization.}
  \label{tab:cross-subject-accuracy}
  \begin{center}
    \begin{normalsize}
        \begin{adjustbox}{max width=\textwidth}
        \renewcommand{\arraystretch}{1.1}
        \setlength{\tabcolsep}{10pt} 
        \begin{tabular}{l@{\hspace{2pt}}c| l l l l}
          \toprule
          \multicolumn{1}{l}{Model} & \multicolumn{1}{c}{UDA} & \multicolumn{1}{c}{Motor Imagery} & \multicolumn{1}{c}{Workload} & \multicolumn{1}{c}{Stress} & \multicolumn{1}{c}{Fatigue} \\
          \midrule
          FBCNet & w/o & 69.5 $\pm$ 10.0 & 49.0 $\pm$ 11.2 & 67.5 $\pm$ 19.7 & 75.2 $\pm$ 8.1 \\
          EEGConformer & w/o & 67.7 $\pm$ 10.4  & 46.5 $\pm$ 11.1  & 72.0 $\pm$ 15.5  & 74.8 $\pm$ 8.5 \\
          ATCNet & w/o & 67.2 $\pm$ 14.9 & 46.6 $\pm$ 9.2  & 66.8 $\pm$ 15.4  & 75.8 $\pm$ 6.3 \\
          EEGDeformer & w/o & 64.5 $\pm$ 8.8 & 48.1 $\pm$ 9.0  & 68.6 $\pm$ 17.4  & 74.7 $\pm$ 10.6 \\
          LaBraM & w/o  & $70.8 \pm 9.7$ & $46.7 \pm 6.8$  & $75.3 \pm 11.7$ & $76.6 \pm 9.9$ \\
          CBraMod & w/o & \textbf{\underline{73.9 $\pm$ 7.3}}  & $49.7 \pm 8.0$  & $79.1 \pm 14.4$ & $77.4 \pm 11.2$ \\
          LUNA  & w/o  & $68.9 \pm 8.6$ & $48.1 \pm 10.3$ & $74.7 \pm 16.2$ & $75.2 \pm 8.9$ \\
          REVE & w/o & 72.5 $\pm$ 8.7   & 48.9 $\pm$ 9.6    & 78.1 $\pm$ 10.2  &  75.7 $\pm$ 11.5\\
          \midrule
          EEGNet & w/o & 66.9 $\pm$ 13.0 & 48.8 $\pm$ 10.7  & 74.9 $\pm$ 19.5 & 75.9 $\pm$ 7.2 \\
           & EA & 73.2 $\pm$ 12.0 & 55.2 $\pm$ 8.6  & 76.9 $\pm$ 18.9 & 75.6 $\pm$ 8.2 \\
           & STMA & 70.6 $\pm$ 9.8 & 56.1 $\pm$ 10.3  & 80.1 $\pm$ 19.2  & 78.3 $\pm$ 8.3 \\
          \midrule
          TSLANet & w/o & 67.1  $\pm$ 13.7 & 50.5 $\pm$ 7.7 & 68.5 $\pm$ 16.7  & 72.2 $\pm$ 7.0  \\
           & EA & 67.2 $\pm$ 12.0  & 54.3 $\pm$ 8.0 & 71.2 $\pm$ 22.6   & 72.9 $\pm$ 10.6 \\
           & STMA & 68.0 $\pm$ 10.7 & 55.7 $\pm$ 8.0 & 74.3 $\pm$ 18.5  & 73.2 $\pm$ 7.9 \\
          \midrule
          SPDMBN   & w/o & 65.5 $\pm$ 11.6  &  47.2 $\pm$ 9.2 & 69.9 $\pm$ 19.2  & 73.4 $\pm$ 10.7  \\
              & RC & 69.0 $\pm$ 14.0  &  56.5 $\pm$ 9.9 & 82.4 $\pm$ 22.4 & 76.6 $\pm$ 12.8  \\
          \midrule
          TSMNet    & SPDDSBN & 70.6 $\pm$ 11.6  &   52.5 $\pm$ 13.0 & 76.4 $\pm$ 20.8  & 78.4 $\pm$ 7.3 \\ 
          HEEGNet & DSMDBN & 73.0 $\pm$ 15.5 & 53.0 $\pm$ 10.9 & 81.5 $\pm$ 15.1 & 76.4 $\pm$ 8.1 \\
          \midrule
            SPDAlign       & - & 72.7 $\pm$ 13.4  &\textbf{\underline{57.6 $\pm$ 10.6}}   & \textbf{\underline{84.5 $\pm$ 18.7}} & \textbf{\underline{82.7 $\pm$ 11.9}} \\
            SPDAlign (Oracle)  & - & 92.6 $\pm$ 5.6 & 88.3 $\pm$ 5.5 & 95.8 $\pm$ 4.3 & 94.8 $\pm$ 5.5 \\
          \bottomrule
        \end{tabular}
        \end{adjustbox}
    \end{normalsize}
  \end{center}
\label{tab_main_exp}
\end{table*}

\subsection{Main experiments} 
We consider four public benchmark EEG datasets: motor imagery \citep{tangermann2012review}, workload estimation \citep{roy_retrospective_2022}, mental stress \citep{zyma2019electroencephalograms}, and driving fatigue \citep{cao2019multi} (see \cref{section_dataset} for detailed descriptions and preprocessing). 
We compare SPDAlign against a comprehensive suite of benchmarks: (i) six deep learning models: EEGNet \citep{lawhern2018eegnet}, EEGConformer \citep{song2022eeg}, ATCNet \citep{altaheri2022physics}, TSLANet \citep{eldele2024tslanet}, EEGDeformer \citep{ding2024eeg}, and FBCNet \citep{mane2021fbcnet}; (ii) four EEG foundation models: LaBraM \citep{jiang2024large}, CBraMod \citep{wangcbramod}, Luna \citep{donerluna} and Reve \citep{elreve}; (iii) two geometric frameworks: SPDMBN \citep{kobler_controlling_2022} and its rotation-corrected variant (SPDMBN+RC); and (iv) UDA frameworks: TSMNet \citep{kobler2022spd}, HEEGNet \citep{li2026heegnet}, Euclidean alignment (EA) \citep{he2019transfer} and spatio-temporal Monge alignment (STMA) \citep{gnassounou2024multi} combined with EEGNet and TSLANet (see \cref{app: implementation details} for detailed implementation).
We evaluate all models under the same evaluation scheme for cross-subject decoding within each dataset.
To establish an empirical upper bound for the Procrustes alignment in the target domain \citep{koc2025domain}, we additionally evaluate a variant of SPDAlign in which target labels are used only during the estimation of the orthogonal transformation through empirical risk minimization (ERM); the rest follows the standard source-ERM protocol.

\begin{figure*}[t]
    \centering
    \includegraphics[width=\textwidth]{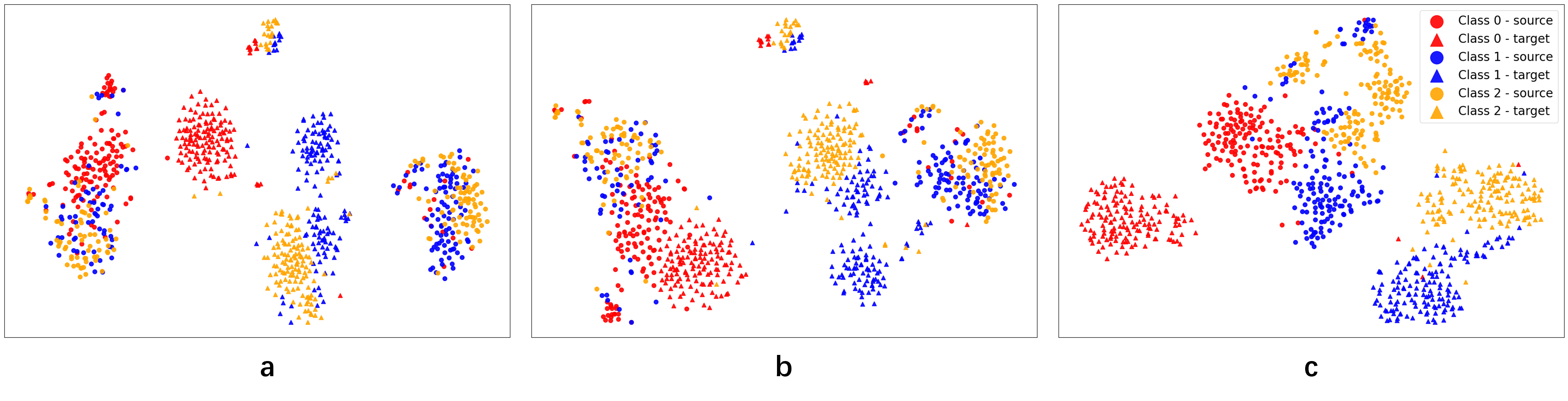}
    \caption{\textbf{SPDAlign steps toward domain-invariant representations.} 
    The t-SNE visualization illustrates the alignment process for the workload dataset in the classification space.
\textbf{(a) Moment alignment:} 
aligning the domain-specific Fréchet means (\cref{eq_dsbn1}) to the identity matrix (previous work).
\textbf{(b) SPDAlign} and \textbf{(c) SPDAlign (Oracle):} applying the orthogonal transformation $Q$ estimated via Wasserstein Procrustes (\cref{eq:wp}) and empirical risk minimization.
SPDAlign enhances class separability, addressing forward modeling shifts without requiring labels.
The remaining gap in the unsupervised setting is primarily due to suboptimal rotation estimation.
}
\label{fig:tsne}
\end{figure*}

\begin{figure*}[t]
    \centering
    \includegraphics[width=\textwidth]{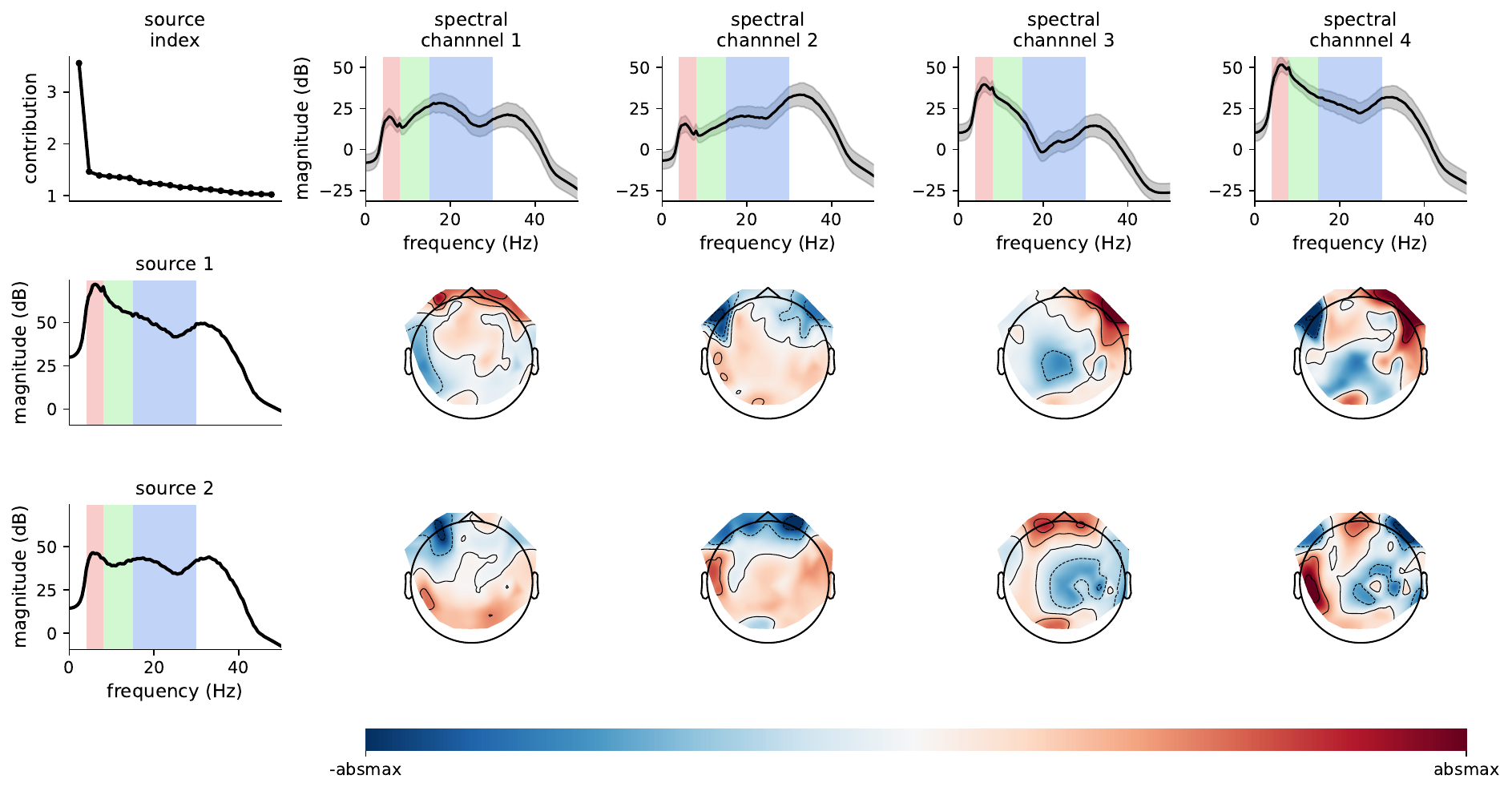}
    \caption{\textbf{SPDAlign interpretability.}
Patterns extracted from the \emph{workload} dataset and class \emph{low} (cross-subject) are shown above. 
The top-left panel details the relative contribution of each extracted latent source. 
For clarity, only the two most discriminative sources are displayed in the left column panels.
The top row panels summarize the frequency profile of each spectral channel (derived from the output of the four temporal convolution layers in $f_{\theta}$). 
Topographic plots show the projection of source activity to regions covered by the EEG channels; areas shaded in darker blue or red indicate greater discriminative source activity. 
These results align with prior research, as the frontal and parietal areas and the theta band serve as biomarkers of mental workload \citep{borghini2014measuring}. 
}
    \label{fig:inter_low}
\end{figure*}

\textbf{Results.} 
\cref{tab_main_exp} summarizes the results across all EEG tasks, presenting the grand average of test-set scores. 
SPDAlign achieves competitive performance, achieving SotA performance in three of four downstream tasks. 
The alignment trajectory for EEG data is visualized using t-SNE in \cref{fig:tsne}.
\cref{fig:inter_low} summarizes SPDAlign interpretation results, where patterns are extracted from the workload dataset and class \emph{low} (cross-subject setting). 
Results for the class \emph{high} can be found in \cref{fig:inter_high}.
For the mental workload tasks, the top contributing source activity peaks in the theta band, and source activities are discriminative primarily in frontal and parietal areas.
These findings are consistent with neuroscience studies identifying that both of these areas and the theta band are relevant markers of workload \citep{borghini2014measuring, dehais2020neuroergonomics}. 
Beyond competitive predictive performance, SPDAlign can identify frequency ranges of interest, determine the specific brain regions (measured by EEG channels), and address cross-subject variability.

\textbf{Model Validation.} 
The high performance of the \emph{Oracle} in \cref{tab_main_exp} supports that forward modeling shifts are the primary cause of distribution shifts in EEG BCI datasets. 
In this setting, the classifier is fixed and shared across domains, while target labels are used only to estimate the optimal rotation for alignment. 
The resulting improvements across datasets suggest that a substantial portion of the cross-subject distribution shift can be addressed through conditional rotation alignment.
These results suggest that the main driver of the domain shift can be explained by transformations in our model, and the remaining gap in the unsupervised setting is primarily due to suboptimal rotation estimation (visualized in \cref{fig:tsne} b and c), not model mismatch.
Correcting for these shifts nearly closes the cross-subject decoding gap, which provides strong empirical validation for the generative model in \cref{eq_forward_model}.
Furthermore, the high performance implies that the underlying latent source covariance is stable across individuals.
These results emphasize the necessity of tackling forward modeling shifts, provide empirical support for our theoretical result, and confirm that the forward model \cref{eq_forward_model} effectively characterizes the physical processes of EEG.

\textbf{Room for improvement.}
While Wasserstein Procrustes estimation allows the unsupervised SPDAlign to achieve state-of-the-art performance, a gap remains relative to the ERM version. 
This discrepancy arises because the Wasserstein Procrustes in \cref{alg:wp} relies on distribution-level optimal transport to infer point correspondences, which can be ambiguous in the absence of label supervision. 
When class-conditional structures overlap or are weakly separated, the estimated transport plan may associate semantically mismatched points, leading to a biased estimation of the global rotation \citep{courty2016optimal}. 
Such erroneous correspondences can accumulate during the iterative optimization and prevent the recovered rotation from reaching the empirical optimum.
Our simulation results (\cref{fig:main_rcc}) further show that unsupervised rotation estimation degrades under extreme rotation shifts or limited class separability. 
These results highlight the potential to develop more robust rotation estimation techniques to bridge the remaining gap toward the oracle upper bound.

\textbf{Ablation.}
\cref{tab_ablation} summarizes the combination of moment and rotation alignment within SPDAlign.
We highlight the three important observations:
(i) the full configuration (both components enabled) achieves the best performance, confirming that both alignment steps are necessary; 
(ii) rotation alignment appears to be particularly critical, as its removal leads to a statistically significant performance drop ($p=0.02$); 
and (iii) the omission of both components results in the largest decrease in accuracy, underscoring their effect in mitigating distribution shifts. 
A global reference mean is used across domains in our implementation; otherwise, the performance will degrade to chance level, which in turn makes it difficult to analyze other parts.
This reduces the apparent impact of explicit moment alignment, as partial moment alignment is already implicitly enforced. 

\begin{table}[h]
\caption{\textbf{Ablation study on the components of SPDAlign.} 
Grand average change in balanced accuracy (\%) for all test sets relative to the full configuration. 
Moment alignment refers to Fr\'echet mean matching, while rotation refers to the (Wasserstein) Procrustes analysis. 
Permutation-paired t-tests, with $t$-values and $p$-values indicating the performance impact of omitting specific components.}
\begin{center}
    \begin{normalsize}
    \begin{adjustbox}{max width=\textwidth}
    \begin{tabular}{cccc}
        \toprule
        moment & rotation  & mean $\pm$ std (\%) & t-val (p-val) \\
        \midrule
        \cmark & \cmark & \multicolumn{1}{c}{--} & \multicolumn{1}{c}{--} \\
        \xmark & \cmark & -2.5 $\pm$ 20.9 & 0.82 (0.4) \\
        \cmark & \xmark & -5.6 $\pm$ 16.3 & 2.37 (0.02) \\
        \xmark & \xmark & -11.3 $\pm$ 18.4 & 3.58 (0.0006) \\
        \bottomrule
    \end{tabular}
    \end{adjustbox}
    \end{normalsize}
\end{center}
    \label{tab_ablation}
\end{table}

\section{Discussions}

In this work, we presented SPDAlign, a global linear Riemannian learning framework for multi-source/multi-target UDA. 
By bridging the physical generative model of EEG with Riemannian geometry, we demonstrated that distribution shifts induced by domain-specific forward modeling can be effectively mitigated through unsupervised alignment on the SPD manifold.
Our results highlight the critical role of rotational alignment; as shown in our ablation study (\cref{tab_ablation}), omitting this step significantly degrades performance.
By achieving SotA results on three of four benchmark datasets (\cref{tab_main_exp}), we show that our approach effectively recovers robust latent representations without requiring target labels.
Furthermore, we demonstrated the interpretability of SPDAlign; the framework can identify frequency ranges of interest and determine the specific brain regions.
Altogether, these results demonstrate the effectiveness of our framework, SPDAlign, which achieves highly competitive performance while retaining inherent interpretability.
The limitation of our framework, which applies to any method involving eigendecompositions, is the computational cost, which limits its application to high-dimensional SPD features. 
While we limited this study to offline evaluations, our framework could be readily extended to online UDA by tracking the running statistics \citep{kobler2022spd} and replacing the mini-batch Wasserstein distance estimation with online variants \citep{mensch2020online}. 
Future work could explore more robust rotation estimation methods and extend the framework to online BCI applications.

\clearpage

\bibliographystyle{iclr2027_conference}
\bibliography{example_paper}

\appendix

\clearpage
\section{SPD manifold in a nutshell}
\label{app: spd_manifold}
The SPD manifold is defined as $\mathcal{S}^{+}_{D} = \{ C \in \mathbb{R}^{D \times D}: C^T = C, C \succ 0 \}$, equipped with an inner product on the tangent space $\mathcal{T}_C\mathcal{S}^{+}_{D}$ at each point $C \in \mathcal{S}^{+}_{D}$,  it forms a Riemannian manifold $M$.
Tangent spaces possess a Euclidean structure, providing a local linear approximation to Riemannian distances on $\mathcal{S}^{+}_{D}$ \citep{absil2008optimization}.
In this study, we adopt the affine-invariant Riemannian metric $g_C^{\text{AIRM}}(S_1, S_2) {=} \mathrm{Tr}(C^{-1}S_1C^{-1}S_2)$ as the chosen inner product, leading to the distance \citep{bhatia2009positive}:
\begin{equation}
\delta(C_1, C_2) = \| \log(C_1^{-\frac{1}{2}} C_2 C_1^{-\frac{1}{2}}) \|_F,
\label{eq_AIRM}
\end{equation}
where $C_1$ and $C_2$ denote two SPD matrices, $\log(\cdot)$ is the matrix logarithm, $\mathrm{Tr}(\cdot)$ is the trace, and $\|\cdot\|_F$ is the Frobenius norm.
Given a set $\mathcal{C} = \{ C_i \in \mathcal{S}^{+}_{D} \}_{i \leq N}$, the Fr\'{e}chet mean is the minimizer of the average squared distances:
\begin{equation}
    \bar{C} = \arg \min_{G \in \mathcal{S}_D^+} \frac{1}{N} \sum_{i=1}^N \delta^2(G, C_i).
\label{eq_FM}
\end{equation}
For $N = 2$, there is a closed-form solution expressed as
\begin{equation}
\bar{C}(\gamma)
=
{C}_1 \#_{\gamma} {C}_2
=
C_1^{\frac{1}{2}}
\left(
C_1^{-\frac{1}{2}}
C_2
C_1^{-\frac{1}{2}}
\right)^{\gamma}
C_1^{\frac{1}{2}}.
\label{eq_fm_2}
\end{equation}
with weight $\gamma = 0.5$. Choosing $\gamma \in [0,1]$ computes weighted means along the geodesic (i.e., the shortest curve) that connects both points. 

The logarithmic map $\text{Log}_{\bar{C}}: \mathcal{S}^{+}_{D} \rightarrow \mathcal{T}_{\bar{C}}\mathcal{S}^{+}_{D}$ projects points from the manifold onto the tangent space at $\bar{C}$ and exponential map $\text{Exp}_{\bar{C}}:\mathcal{T}_{\bar{C}}\mathcal{S}^{+}_{D}\ \rightarrow \mathcal{S}^{+}_{D}$ performs the inverse projection.
\begin{align}
\operatorname{Log}_{\bar{C}}(C_i)
&= \bar{C}^{\frac{1}{2}}
\log\!\left(
\bar{C}^{-\frac{1}{2}} C_i \bar{C}^{-\frac{1}{2}}
\right)
\bar{C}^{\frac{1}{2}} ,
\label{eq_log_map}\\
\text{Exp}_{\bar{C}}(S_i) &= {\bar{C}}^{\frac{1}{2}} \exp({\bar{C}}^{-\frac{1}{2}} S_i {\bar{C}}^{-\frac{1}{2}}) {\bar{C}}^{\frac{1}{2}}.
\label{eq_exp_map}
\end{align}
Parallel transport $\Gamma_{\bar{C} \to \bar{C_{\phi}}} \mathcal{T}_{\bar{C}}S_{D}^{+} \to \mathcal{T}_{\bar{C}_{\phi}}S_{D}^{+}$ transports tangent space vectors $S_i \in \mathcal{T}_{\bar{C}}\mathcal{S}^{+}_{D}$ at $\bar{C}$ to the tangent space of an arbitrary point $\bar{C}_{\phi}$ on the manifold:
\begin{equation}
\Gamma_{\bar{C} \rightarrow \bar{C}_{\phi}} (S_i)
= \left( \bar{C}_{\phi} \bar{C}^{-1} \right)^{\frac{1}{2}} \, S_i \, \left( \bar{C}_{\phi} \bar{C}^{-1} \right)^{\frac{1}{2}}.
\label{eq_paratransport}
\end{equation}

\subsection{Related work}
\label{app_rw}
Riemannian geometry-aware alignment frameworks have dominated in EEG cross-domain generalization problems \citep{roy_retrospective_2022, ju2025spd, mellot2023harmonizing}.
Within the typical sequential TSM learning framework, \citep{rodrigues_riemannian_2019, bleuze2022tangent} match a source and a target distribution using simple geometric transformations: first- and second-order moment alignment and a global rotation transformation over the data points in a semi-supervised fashion.
More recently, \citep{kobler2022spd} proposed tracking the domain-specific running Fr\'{e}chet mean during training, thereby achieving end-to-end learning of feature extractors, TSM, and a tangent space classifier.

While our framework adopts similar geometric transformations, it introduces several key distinctions.
Unlike the previous end-to-end learning framework of \citep{kobler2022spd}, which aligns only domain-specific moments, SPDAlign additionally accounts for rotation transformations. 
We further show the importance of rotation alignment in multi-source, multi-target settings that was overlooked in the framework in \citep{kobler2022spd} through simulations (\cref{fig:domain}) and ablation (\cref{tab_ablation}). 
In contrast to existing rotation alignment methods \citep{rodrigues_riemannian_2019, bleuze2022tangent} that rely on target domain labels for estimation, SPDAlign is the first unsupervised framework capable of estimating these rotations in EEG without label access. 
Additionally, SPDAlign addresses the multi-source, multi-target adaptation setting, reflecting real-world BCI scenarios where data from multiple recorded subjects are leveraged to enable continuous adaptation to new users \citep{he2020different}.

\clearpage

\section{Derivation of domain-invariant latent representations}
\label{appendix_prop_conditional_shifts}

Leveraging the linear relationship between the latent covariance and the empirical covariance (\cref{eq_model_C_i}), together with the congruence invariance property of the Fr\'{e}chet mean \citep{bhatia_riemannian_2013}, we obtain:
\begin{equation}
    \bar{C}_j  = A_j \bar{E}_j A_j^T
    \label{eq_fm_congruence_invariance}
\end{equation}
where $\bar{E}_j$ is the Fr\'{e}chet mean of the latent covariance.

In previous work, \cite{lispdim} prove a proposition that for the set of examples $\mathcal{E}_i = \{ E_i  | E_i \in \mathcal{S}_P^+\}_{i=1}^{L}$, the Fr\'{e}chet mean $\mathcal{E}_j$, defined in (\ref{eq_FM}), converges as $L \rightarrow \infty$ to the identity matrix $I_P$ (i.e., $\bar{E}_j = I_P$ for all $j$).
Using the proposition $\bar{E}_j = I_P$ for all $j$ and substituting (\cref{eq_pd}) for $A_j$, the expression in (\cref{eq_fm_congruence_invariance}) simplifies to:
\begin{align}
    \bar{C}_j &= A_j I_P A_j^T \\
              &= Q_j \exp(P_j) \exp(P_j) Q_j^T \\
              &= Q_j \exp(P_j)^2 Q_j^T \\
    \Rightarrow \bar{C}_j^{-\frac{1}{2}} &= Q_j \exp(P_j)^{-1} Q_j^T
    \label{eq_c_mean}
\end{align}

The two-stage transformations, as defined in \cref{eq_dsbn1} and \cref{eq_dsbn}, consisting of recentering the data $C_i \in \mathcal{C}_j$ and a global rotation transformation for each domain $j$, together with \cref{eq_c_mean}, \cref{eq_pd}, and \cref{eq_paratransport} gives:
\begin{align}
    m_\phi(C_{ij}) &= Q_{j}^T(\Gamma_{\bar{C}_j \rightarrow I} \circ \mathrm{Log}_{\bar{C}_j}(C_{ij})) Q_{j}\\
    &= Q_{j}^T(\Gamma_{\bar{C}_j \rightarrow I} \circ(\bar{C}_j^{\frac{1}{2}}
       \log\!\left(\bar{C_j}^{-\frac{1}{2}} C_{ij} \bar{C_j}^{-\frac{1}{2}}\right)
        \bar{C}_j^{\frac{1}{2}}))Q_{j}\\
    &= Q_{j}^T(\log\left( \bar{C}_j^{-\frac{1}{2}}\, C_{ij}\, \bar{C}_j^{-\frac{1}{2}} \right))Q_{j}\\
    &= Q_{j}^T\log\left( Q_j \exp(P_j)^{-1} Q_j^T C_{ij} Q_j \exp(P_j)^{-1} Q_j^T \right)Q_{j} \\
    &= Q_{j}^T\log\left( Q_j \exp(P_j)^{-1} A_j E_i A_j^T \exp(P_j)^{-1} Q_j^T \right)Q_{j} \\
    &= Q_{j}^T\log\left( Q_j \exp(P_j)^{-1} Q_j^T Q_j \exp(P_j) E_i \exp(P_j) Q_j^T Q_j \exp(P_j)^{-1} Q_j^T \right)Q_{j} \\
    &= Q_{j}^T\log\left( Q_j E_i  Q_j^T \right)Q_{j} \\
    &= Q_{j}^T Q_j \log(E_i) Q_j^T Q_{j} \\
    &= \log(E_i)
\end{align}
where \cref{eq_model_C_i} is used for $C_{ij}$, for any invertible matrix $G$ and any $X \in \mathcal{S}_P^+$, the matrix logarithm satisfies
$\log(G X G^{-1}) = G \log(X) G^{-1}$.

\clearpage

\section{Simulations}
\label{app_section_simulation}
To construct the covariance matrices \( C_i \in \mathcal{S}_P^+ \) with \( P=2 \), we initiated the generation process in the log-space. 
Specifically, log-space features, denoted as \( \mathrm{upper} \log (E_i) \), were synthesized using the \texttt{scikit-learn} function \texttt{make\_classification}, with one dimension configured to encode label-relevant information. 
These features were subsequently standardized to achieve zero mean and unit variance. 
To map these points onto the SPD manifold \( \mathcal{S}_P^+ \), we applied the inverse vectorization operator \( \mathrm{upper}^{-1} \) and the exponential map \( \mathrm{Exp}_{I_P} \) as defined in (\ref{eq_exp_map}). 
The resulting data were partitioned across source and target domains, with each domain receiving 500 observations.
Finally, the data were projected into the channel space via domain-specific forward matrices \( A_j \), following (\ref{eq_pd}). 
To introduce rotation shifts, we either artificially varied the rotation degree using angle-parameterized rotation matrices or generated random orthogonal matrices.

\begin{figure*}[h]
    \centering
    \includegraphics[width=\textwidth]{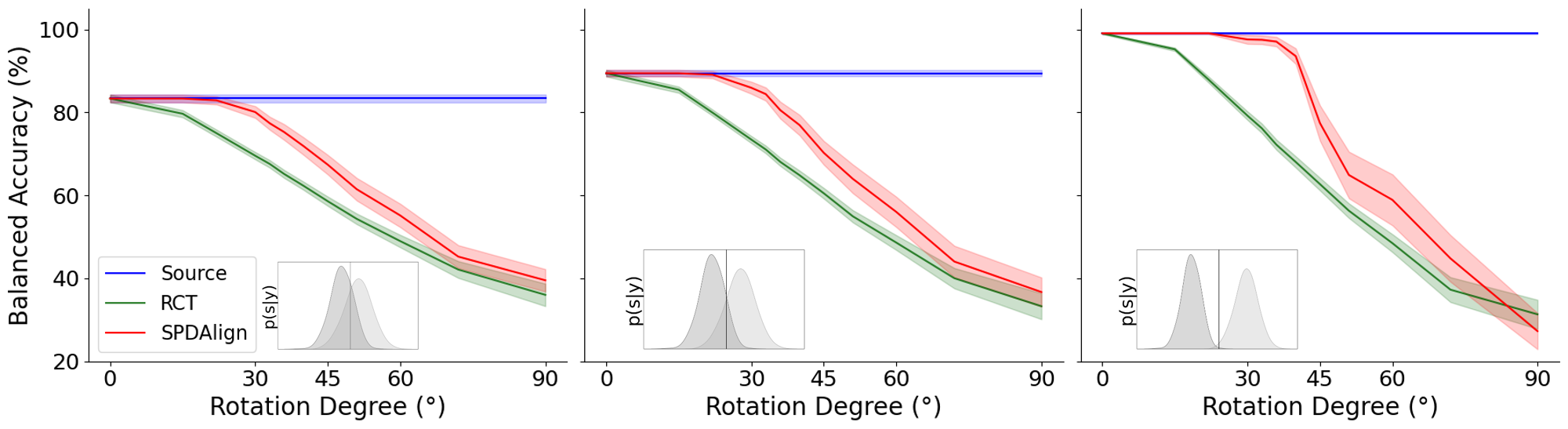}
    \caption{\textbf{Simulation performance evaluation across varying class separability.} 
    The panels illustrate the balanced accuracy ($y$-axis) of different alignment methods as a function of the rotational shift in degrees ($x$-axis) introduced by domain-specific forward modeling shifts. 
    Each panel represents a distinct level of class separability, with the degree of class overlap visualized in the gray insets.
    We compare the performance of SPDAlign against recentering to the identity (RCT) \citep{zanini2017transfer}. 
    The blue line indicates source domain performance, providing an empirical upper bound for the UDA task.
    }
    \label{fig:main_rcc}
\end{figure*}

Class separability \cref{fig:main_rcc}: in the absence of rotation, moment alignment alone effectively compensates for the domain-specific transition effects introduced by EEG forward modeling shifts (\cref{eq_pd}).
As expected, classification performance declines as the magnitude of rotational shifts increases across all scenarios, though higher class separability confers greater robustness.
Notably, while the RCT method (green) is highly sensitive to even mild rotations, SPDAlign (red) demonstrates superior resilience, maintaining stable accuracy across a wider range of shifts.
The source-domain performance (blue line) represents the empirical upper bound for these unsupervised adaptation tasks.

\begin{figure*}[h]
    \centering
    \includegraphics[width=\textwidth]{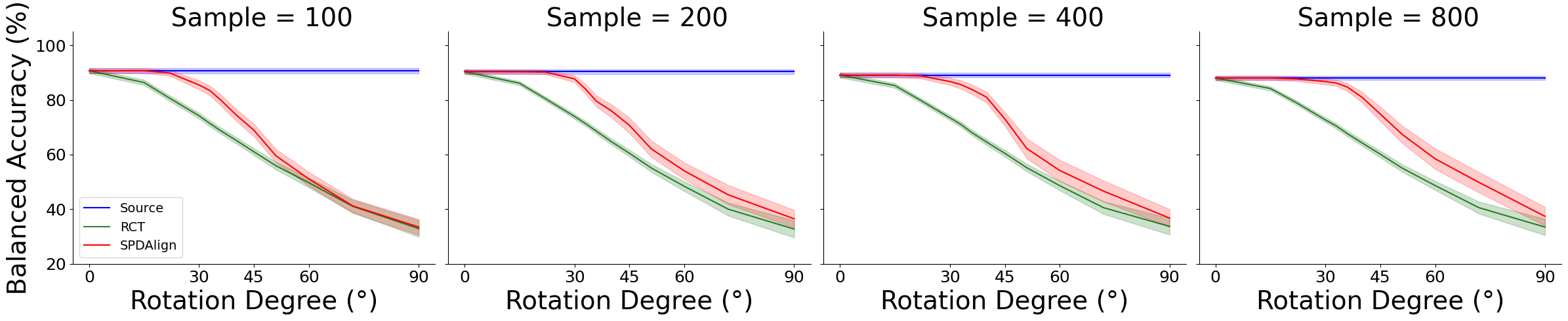}
    \caption{\textbf{Simulation performance over the number of data samples.} Balanced accuracy scores (higher is better) across rotation degree on the x-axis and number of data samples across panels. 
    Larger sample sizes lead to a clearer advantage of SPDAlign over RCT under moderate rotational shifts.}
    \label{fig:app_sam}
\end{figure*}

Sample size \cref{fig:app_sam}: Performance degrades gradually as the number of samples decreases, primarily affecting the estimation of covariance matrices and Wasserstein Procrustes.
However, SPDAlign maintains a clear advantage once moderate sample sizes are available (e.g., typical EEG epoch counts), even when the rotation is severe.

\begin{figure*}[h]
    \centering
    \includegraphics[width=\textwidth]{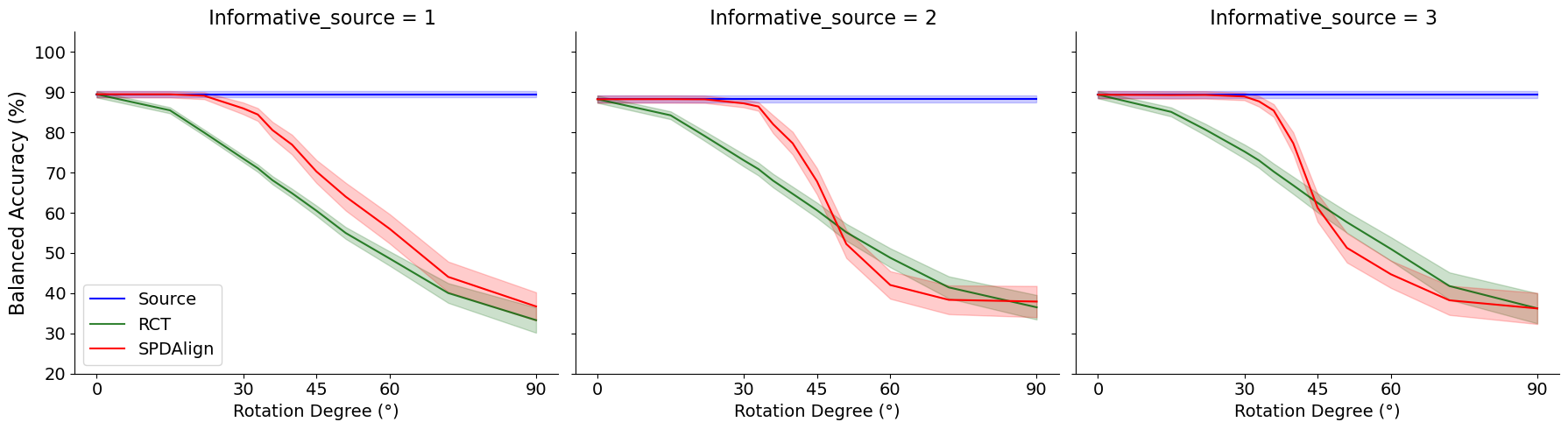}
    \caption{\textbf{Simulation performance over the number of informative sources.} Balanced accuracy scores (higher is better) across rotation degree on the x-axis and number of informative sources across panels.
     SPDAlign generally remains more robust than RCT; the advantage is particularly evident when sufficient informative source structure is available, supporting the robustness of SPDAlign under typical EEG regimes.}
    \label{fig:app_info}
\end{figure*}

Informative sources (\cref{fig:app_info}): we vary the number of label-informative latent sources to evaluate the effect of distributed task-relevant information.
Across different settings, classification performance decreases as rotational shifts become more severe, while SPDAlign generally remains more robust than RCT. 
The advantage is particularly evident when sufficient informative source structure is available, supporting the robustness of SPDAlign under typical EEG regimes.

\begin{figure*}
    \centering
    \includegraphics[width=\textwidth]{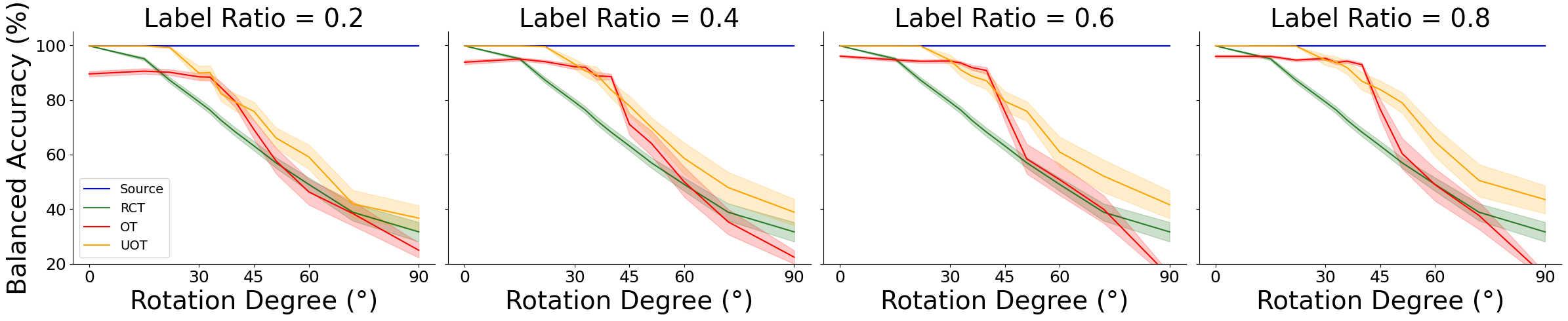}
    \caption{\textbf{Simulation performance over label shift ratios.} Balanced accuracy scores (higher is better) across rotation degree on the x-axis.
    As expected, label shifts degraded performance, but UOT, which relaxes the marginal constraints, still provided performance gains, indicating that UOT is a suitable choice when class imbalance is present.}
    \label{fig:app label shift}
\end{figure*}

Label shifts (\cref{fig:app label shift}): the ratio represents the degree of class imbalance in the simulation. 
We compared the optimal transport (OT) method used in our paper with its unbalanced variant, UOT. 
Performance was tested under four different ratios, ranging from mild to severe label shifts. 
As expected, label shifts degraded performance, but UOT, which relaxes the marginal constraints, still provided performance gains, indicating that UOT is a suitable choice when class imbalance is present.

\begin{figure*}
    \centering
    \includegraphics[width=\textwidth]{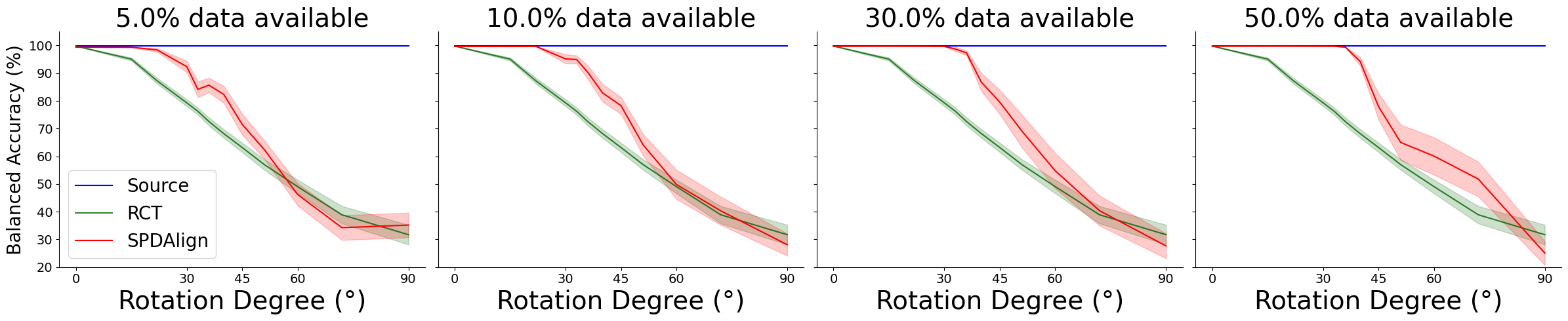}
    \caption{\textbf{Simulation performance over available data percentage.} Balanced accuracy scores (higher is better) across rotation degree on the x-axis.
    With limited incoming data, the estimation is less reliable, resulting in a smaller advantage over RCT. 
    As more target samples become available, SPDAlign consistently shows advantages over RCT.}
    \label{fig:app data percentage}
\end{figure*}

Data availability (\cref{fig:app data percentage}): to examine the effect of data availability in an online learning setting, we progressively vary the proportion of target-domain samples available for alignment. 
With limited incoming data, the estimation is less reliable, resulting in a smaller advantage over RCT. 
As more target samples become available, SPDAlign consistently shows advantages over RCT. 
This result suggests that SPDAlign can be easily extended to online settings, where its alignment quality progressively improves as data accumulate.

\begin{figure}[h]
    \centering
    \includegraphics[width=0.48\textwidth]{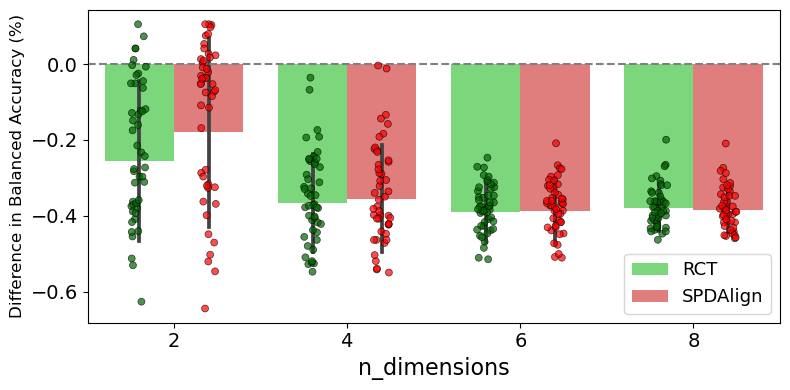}
    \caption{\textbf{Simulation performance over the number of data dimensions.} Balanced accuracy scores (higher is better) across the number of data dimensions on the x-axis.
    As the proportion of noise sources increases (i.e., as SNR decreases), all methods’ performance declines.
    SPDAlign shows a clear advantage under mild noise conditions, remains beneficial under moderate noise, and maintains similar performance
    without degrading results under more extreme noise levels.
    }
    \label{fig:dim}
\end{figure}

SNR (\cref{fig:dim}): As SNR decreases (i.e., increasing proportion of noise sources in the simulation), all methods' performance declines.
SPDAlign shows a clear advantage under mild noise conditions (-3 dB), remains beneficial under moderate noise (-10.5 dB), and maintains similar performance without degrading results under more extreme noise levels (-14.6 dB, -17.4 dB).

\begin{figure}[h]
    \centering
    \includegraphics[width=0.48\textwidth]{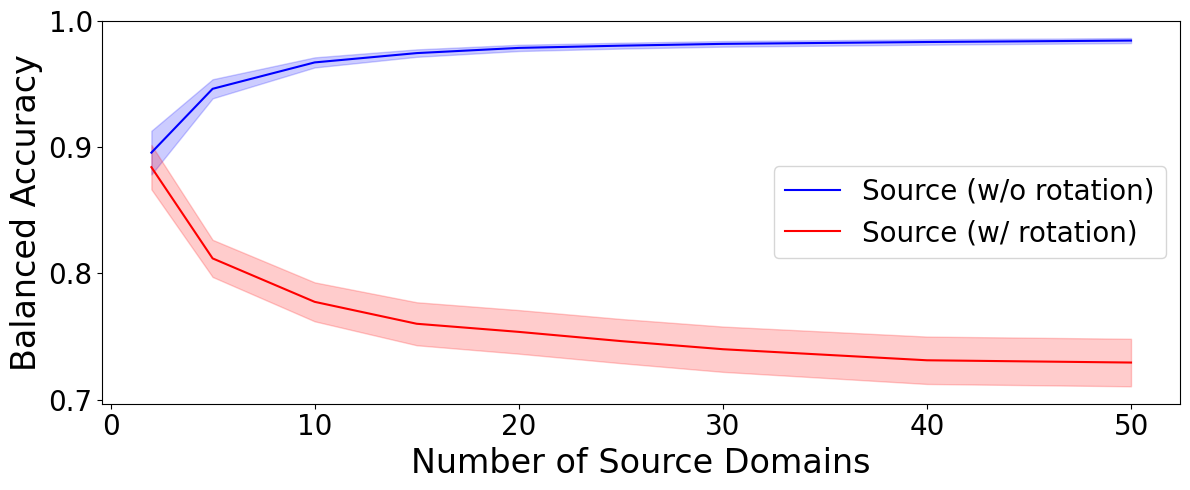}
    \caption{\textbf{Simulation performance over the number of source domains.} Balanced accuracy scores (higher is better) across the number of source domains on the x-axis.
    In the absence of source domain rotation correction, training performance degrades as the number of source domains increases, due to the accumulation of conflicting domain-specific rotation shifts.}
    \label{fig:domain}
\end{figure}

Domain numbers (\cref{fig:domain}): we evaluate performance on consistent target domains across a varying number of source domains, comparing the impact of enabling or disabling source domain rotation correction. 
In the absence of such correction, training performance degrades as the number of source domains increases, due to the accumulation of conflicting domain-specific rotation shifts, underscoring the importance of source-domain alignment in multi-source settings that was overlooked in the framework proposed by \cite{kobler2022spd}.

\clearpage

\section{Experiment details}

\subsection{Dataset details}
\label{section_dataset}
\textbf{Motor imagery}:
\citep{tangermann2012review} is a motor imagery dataset containing EEG from 9 subjects recorded with 22 channels at 250 Hz. 
The paradigm involves two motor imagery tasks: left hand and right hand, with 144 trials per class per subject. 
Data were collected across two sessions on separate days, each comprising six runs of 48 trials. 
Each trial lasted 4 seconds.  

\textbf{Workload}: \citep{roy_retrospective_2022} containing EEG from 15 subjects recorded with 64 channels at 500 Hz. The paradigm involves three distinct mental workload levels: low, medium, and high, induced by the Multi-Attribute Task Battery II (MATB-II) task. 
Each participant completed three sessions on different days, with each workload condition lasting 5 minutes per session.

\textbf{Mental Stress}: \citep{zyma2019electroencephalograms} containing EEG from 36 subjects recorded with 19 channels at 500 Hz using the Neurocom system. 
The paradigm involves two cognitive states: a resting state and a mental stress state induced by an intensive serial subtraction arithmetic task. 
Each participant completed a single session consisting of a 3-minute resting period followed by 4 minutes of mental arithmetic.

\textbf{Driving Fatigue}:
\citep{cao2019multi} containing EEG from 11 subjects recorded with 32 channels at 500 Hz. 
The paradigm involves a prolonged 90-minute dynamic driving simulation designed to induce varying levels of fatigue. 
Continuous EEG recordings were collected, with lane departure events and reaction times monitored to assess alertness and fatigue throughout the session.

\textbf{Preprocessing.}
For motor imagery, signals were resampled to 256 Hz, band-pass filtered between 4–36 Hz, and segmented into 3-second epochs corresponding to the two most discriminative labels. 
For workload, signals were band-pass filtered at 4–36 Hz, resampled to 200 Hz, and segmented into 2-second epochs. 
For driving fatigue, signals were band-pass filtered at 1–50 Hz, resampled to 128 Hz, and segmented into 3-second epochs. 
For mental stress, signals were processed with a 30 Hz low-pass filter and a 50 Hz notch filter, resampled to 500 Hz, and segmented into 4-second epochs. 

\subsection{Implementation details}
\label{app: implementation details}

All models were implemented using official repositories, maintaining the default hyperparameters.
Each session is treated as a domain. 
We evaluate using either leave-one-subject-out (subject count $\leq 10$) or 10-fold cross-validation, with subjects serving as the grouping variable for cross-subject adaptation. 
Euclidean frameworks use the Adam optimizer \citep{kingma2014adam}, while geometric frameworks use Riemannian Adam \citep{becigneul2018riemannian}, both with default PyTorch hyperparameters. 
Source-domain data are split into 80\% training and 20\% validation sets, stratified by domain and label, and trained for 50 epochs with cross-entropy loss. 

Baseline methods were implemented using publicly available Python code; custom approaches were developed using PyTorch \citep{paszke2019pytorch}, scikit-learn \citep{pedregosa2011scikit}, and Geoopt \citep{geoopt2020kochurov}. 
All experiments were performed on standard computation PCs equipped with 32-core CPUs, 128 GB of RAM, and a single GPU.

\clearpage
\section{SPDAlign Details}
\label{sec_spdalign_architecture}
\subsection{Model parameters}

For the feature extractor $ f_{\theta}$, we adopt the architecture from \citet{kobler2022spd}, which includes two linear convolutional layers, covariance pooling \citep{acharya2018covariance}, and BiMap and ReEig layers \citep{huang_riemannian_2017}.
Specifically, the BiMap layer performs a linear subspace projection, and the ReEig layer thresholds the eigenvalues of symmetric matrices with the default threshold of $10^{-4}$; it was never activated.

\begin{table}[h]
\centering
\small
\renewcommand{\arraystretch}{1.2}
\caption{SPDAlign parameters}
\begin{adjustbox}{max width=\textwidth}
\begin{tabular}{lll}
\toprule
Block & Output (dim) & Parameter (dim) \\
\midrule
TempConv  & $4 \times P \times T$   & $4 \times 1 \times 1 \times 25$ \\
SpatConv  & $40 \times 1 \times T$  & $40 \times 4 \times P \times 1$ \\
CovPool   & $40 \times 40$          & --                              \\
BiMap     & $20 \times 20$          & $40 \times 20$                  \\
ReEig     & $20 \times 20$          & --                              \\
SPDAlign  & $20 \times 20$          & $1$                             \\
LogEig    & $210$                   & --                              \\
Linear    & $C$                     & $211 \times C$                  \\
\bottomrule
\end{tabular}
\end{adjustbox}
\end{table}

\subsection{SPDAlign source domain training}

\begin{algorithm}[h]
\caption{SPDAlign source training}
\label{alg:training}
\begin{algorithmic}[1]

\STATE {\bfseries Input:}
batch
$\mathcal{B}_k
=
\{(\mathbf{x}_i,y_i,d(i))\}_{i=1}^{M}$,
where $d(i)$ denotes the associated source domain;
domain-specific running Fr\'echet means
$\{\bar{\mathbf C}_{k-1}^{d}\}_{d\in\mathcal J_s}$,
with $\bar{\mathbf C}_{0}^{d}=\mathbf I$;
training momentum $\eta_k$;
model parameters: feature extractor $f_{\theta}$ and
classifier $g_{\psi}$

\STATE {\bfseries Output:}
normalized representations
$\{\widetilde{\mathbf Z}_i\}$,
updated running means
$\{\bar{\mathbf C}_{k}^{d}\}$,
and updated $\theta,\psi$

\vspace{0.2em}

\STATE $\mathbf C_i = f_{\theta}(\mathbf x_i)$
\COMMENT{extract SPD features}

\FOR{each domain $d$ in $\mathcal B_k$}

    \STATE
    $\displaystyle
    \mathbf G_k^{d}
    =
    \arg\min_{\mathbf G\in\mathcal S_D^{+}}
    \frac{1}{|\mathcal B_k^{d}|}
    \sum_{i:d(i)=d}
    \delta^2(\mathbf G,\mathbf C_i)$
    \COMMENT{domain-specific batch mean, \cref{eq_FM}}

    \STATE
    $\bar{\mathbf C}_{k}^{d}
    =
    \bar{\mathbf C}_{k-1}^{d}
    \#_{\eta_k}
    \mathbf G_k^{d}$
    \COMMENT{update running mean, \cref{eq_fm_2}}

    \STATE
    $\mathbf Z_i
    =
    \log\!\left[
    \left(\bar{\mathbf C}_{k}^{d}\right)^{-\frac12}
    \mathbf C_i
    \left(\bar{\mathbf C}_{k}^{d}\right)^{-\frac12}
    \right],
    \quad d(i)=d$
    \COMMENT{domain-specific recentering, \cref{eq_dsbn1}}

\ENDFOR

\STATE
$\hat y_i
=
g_{\psi}
\left(
\operatorname{upper}(\mathbf Z_i)
\right)$

\STATE
Update $\theta$ and $\psi$ by minimizing
$\sum_i \ell(\hat y_i,y_i)$

\vspace{0.2em}

\STATE {\bfseries Rotation alignment stage:}

\STATE Freeze $\theta$ and randomly select
reference domain $r\in\mathcal J_s$

\FOR{each source domain $d\neq r$}

    \STATE Construct class-constrained correspondences
    between domains $d$ and $r$

    \STATE Downsample unequal class-wise sample counts

    \STATE
    $\widehat{\mathbf Q}^{d}
    =
    \operatorname{Procrustes}
    (\mathbf Z^{d},\mathbf Z^{r})$
    \COMMENT{estimate source rotation, \cref{eq:procrustes_solution}}

    \STATE
    $\widetilde{\mathbf Z}_i
    =
    (\widehat{\mathbf Q}^{d})^{\top}
    \mathbf Z_i
    \widehat{\mathbf Q}^{d}
     = \log(E_i),
    \quad d(i)=d$
    \COMMENT{normalized representations, \cref{eq_dsbn}}

\ENDFOR

\STATE
$\widehat{\mathbf Q}^{r}=\mathbf I$

\STATE
Fine-tune $\psi$ using
$\{\widetilde{\mathbf Z}_i,y_i\}$,
while keeping $\theta$ fixed

\end{algorithmic}
\end{algorithm}

\clearpage
\subsection{Wasserstein Procrustes in SPDAlign}

\begin{algorithm}[h]
\caption{Wasserstein Procrustes in SPDAlign}
\label{alg:wp}
\begin{algorithmic}[1]
\STATE {\bfseries Input:} Source data $\mathbf{X} \in \mathbb{R}^{n_a \times d}$, Target data $\mathbf{Y} \in \mathbb{R}^{n_b \times d}$, learning rate $\alpha$, iterations $T$
\STATE Initialize $\mathbf{Q}_0 \in \mathcal{O}_d$ (orthogonal matrix)
\STATE Set uniform weights $\mathbf{a} = \frac{1}{n_a} \mathbf{1}$, $\mathbf{b} = \frac{1}{n_b} \mathbf{1}$

\FOR{$t = 1$ {\bfseries to} $T$}
    \STATE Compute the optimal transport coupling given $\mathbf{Q}_t$:
    \[
    \boldsymbol{\pi}_t
    =
    \arg\min_{\boldsymbol{\pi} \in \Pi(\mathbf a,\mathbf b)}
    \sum_{i,j} \boldsymbol{\pi}_{ij}
    \left\|
    \mathbf Q_t \mathbf x_i - \mathbf y_j
    \right\|_2^2
    \]
    
    \STATE Compute the gradient with respect to $\mathbf{Q}$:
    \[
    \mathbf{G}_t = -2 \mathbf{X}^\top \boldsymbol{\pi}_t \mathbf{Y}
    \]
    
    \STATE Projected Gradient Descent on the orthogonal group:
    \[
    \mathbf{Q}_{t+1} = \Pi_{\mathcal{O}_d}\!\left( \mathbf{Q}_t - \alpha \mathbf{G}_t \right)
    \]
    
\ENDFOR
\STATE The projection $\Pi_{\mathcal{O}_d}(\mathbf{Q})$ is given by $\mathbf{U} \mathbf{V}^\top$,  where $\mathbf{U} \boldsymbol{\Sigma} \mathbf{V}^\top$ is the SVD of $\mathbf{Q}$.
\STATE {\bfseries Output:} $\mathbf{Q} = \mathbf{U} \mathbf{V}^\mathrm{T}$
\end{algorithmic}
\end{algorithm}

\subsection{Interpretability}

\begin{figure*}[h]
    \centering
    \includegraphics[width=\textwidth]{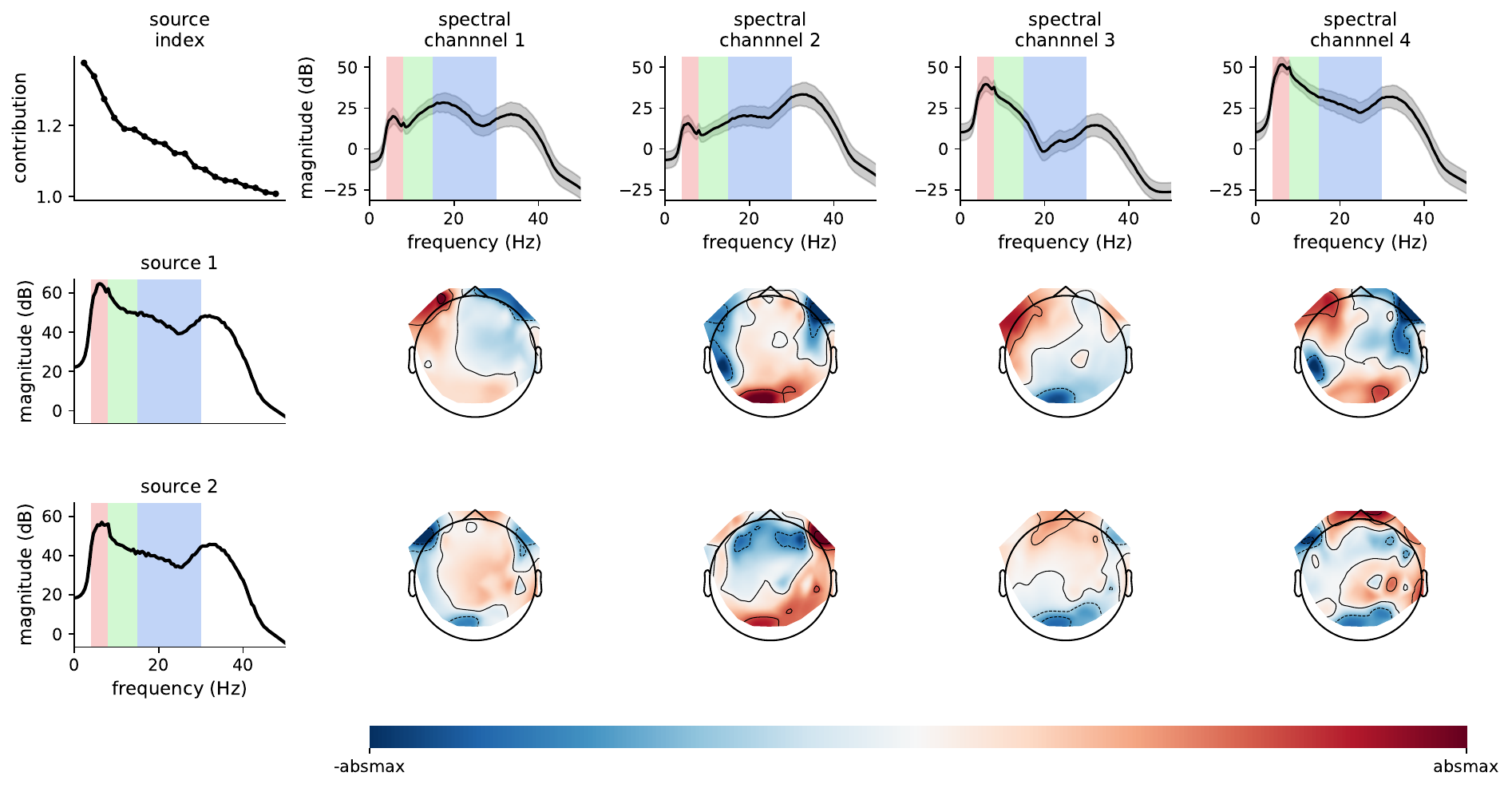}
    \caption{SPDAlign interpretation results. Patterns extracted from the workload dataset and class high
(cross-subject setting). }
    \label{fig:inter_high}
\end{figure*}

\clearpage

\end{document}